\documentclass[runningheads]{llncs}
\usepackage{graphicx}
\usepackage[misc]{ifsym}
\usepackage[hyphens]{url}
\usepackage{hyperref}
\hypersetup{breaklinks=true}

\usepackage[table]{xcolor}

\usepackage{epstopdf}
\usepackage{subfigure}

\usepackage{nameref}
\usepackage{amsmath}

\usepackage{amssymb}

\usepackage{comment}
\usepackage{bm}

\DeclareMathOperator{\census}{census}
\DeclareMathOperator{\pixels}{pixels}
\DeclareMathOperator{\units}{units}
\DeclareMathOperator{\surface}{surface}
\DeclareMathOperator{\prediction}{prediction}

\DeclareMathOperator*{\argmin}{arg\,min}
\usepackage{todonotes}
\usepackage{tabularx}

\usepackage{array}
\newcolumntype{L}[1]{>{\raggedright\let\newline\\\arraybackslash\hspace{0pt}}m{#1}}
\newcolumntype{C}[1]{>{\centering\let\newline\\\arraybackslash\hspace{0pt}}m{#1}}
\newcolumntype{R}[1]{>{\raggedleft\let\newline\\\arraybackslash\hspace{0pt}}m{#1}}

\usepackage{multirow}
\usepackage{booktabs}

\newcolumntype{+}{!{\vrule width 2pt}}

\newlength\savedwidth

\newcommand\thickhline{\noalign{\global\savedwidth\arrayrulewidth\global\arrayrulewidth 2pt}%
	\hline
	\noalign{\global\arrayrulewidth\savedwidth}}

\toctitle{An aggregate learning approach for interpretable semi-supervised population prediction and disaggregation using ancillary data}
\tocauthor{Guillaume Derval,  Fr\'ed\'eric Docquier and Pierre Schaus}

\begin{document}
\title{An aggregate learning approach for interpretable semi-supervised population prediction and disaggregation using ancillary data}
\titlerunning{Interpretable semi-supervised population prediction and disaggregation}
%
\author{Guillaume Derval\inst{1}\orcidID{0000-0002-6700-3519} {\Letter}\and
Fr\'ed\'eric Docquier\inst{2} \and
Pierre Schaus\inst{1}\orcidID{0000-0002-3153-8941}}
\authorrunning{G. Derval et al.}

\institute{ICTEAM, UCLouvain, Louvain-la-Neuve, Belgium \and
IRES, UCLouvain, Louvain-la-Neuve, Belgium\\
\email{\{first\}.\{last\}@uclouvain.be}}
\maketitle              
\begin{abstract}
Census data provide detailed information about population characteristics at a coarse resolution. Nevertheless, fine-grained, high-resolution mappings of population counts are increasingly needed to characterize population dynamics and to assess the consequences of climate shocks, natural disasters, investments in infrastructure, development policies, etc. Dissagregating these census is a complex machine learning, and multiple solutions have been proposed in past research. We propose in this paper to view the problem in the context of the aggregate learning paradigm, where the output value for all training points is not known, but where it is only known for aggregates of the points (i.e. in this context, for regions of pixels where a census is available). We demonstrate with a very simple and interpretable model that this method is on par, and even outperforms on some metrics, the state-of-the-art, despite its simplicity.

\end{abstract}

\section{Introduction}
Most countries periodically organize rounds of censuses of their population at a granularity that differs from country to country. The level of disaggregation is often governed by the administrative division of the country. Although census data are usually considered as accurate in terms of population counts and characteristics, the spatial granularity, that is sometimes in the order of hundreds of square kilometers, is too coarse for evaluating local policy reforms or for making informed decisions about health and well-being of people, economic and environmental interventions, security, etc. (see \cite{UNResolution}). For example, fine-grained, high-resolution mappings of the distribution of the population are required to assess the number of people living at or near sea level, near hospitals, in the vicinity or airports and highways, in conflict areas, etc. They are also needed to understand how population movements react to various types of shocks such as natural disasters, conflicts, plant creation and closures, etc.

Multiple methods can be used to produce gridded data sets (also called rasters), with pixels of a relatively small scale compared to the administrative units of the countries. Notably, the Gridded Population of the World (GPW) \cite{GPW} provides a gridded dataset of the whole world, with a resolution of 30 arcseconds (i.e. pixels of approximately 1 km$^2$ at the equator). The method used in GPW assumes that population is uniformly distributed within each administrative unit. With minor adjustments for boundaries and water bodies, the density of the population in a pixel is identical to the density of the population of the administrative unit in the underlying census. 

More advanced and successful models rely on ancillary and remotely sensed data, and are trained using machine learning techniques (see \cite{ll,rf,robinson2017deep}). These data can include information sensed by satellite (nighttime light intensities, temperature, etc.) or data provided by NGOs and governments (on health facilities, road infrastructure, etc.). These models are used to \textit{disaggregate} coarse census data into a grid of small pixels, using ancillary data to predict the population distribution within the census unit. The level of resolution is either imposed by the method or decided by the user. As an example, such models predict that people concentrate near the road infrastructure rather than in cultivation fields; or that the density of the population in highly elevated regions is smaller than near the sea level.

\begin{figure}[ht!]
	\centering
	\includegraphics[width=\textwidth]{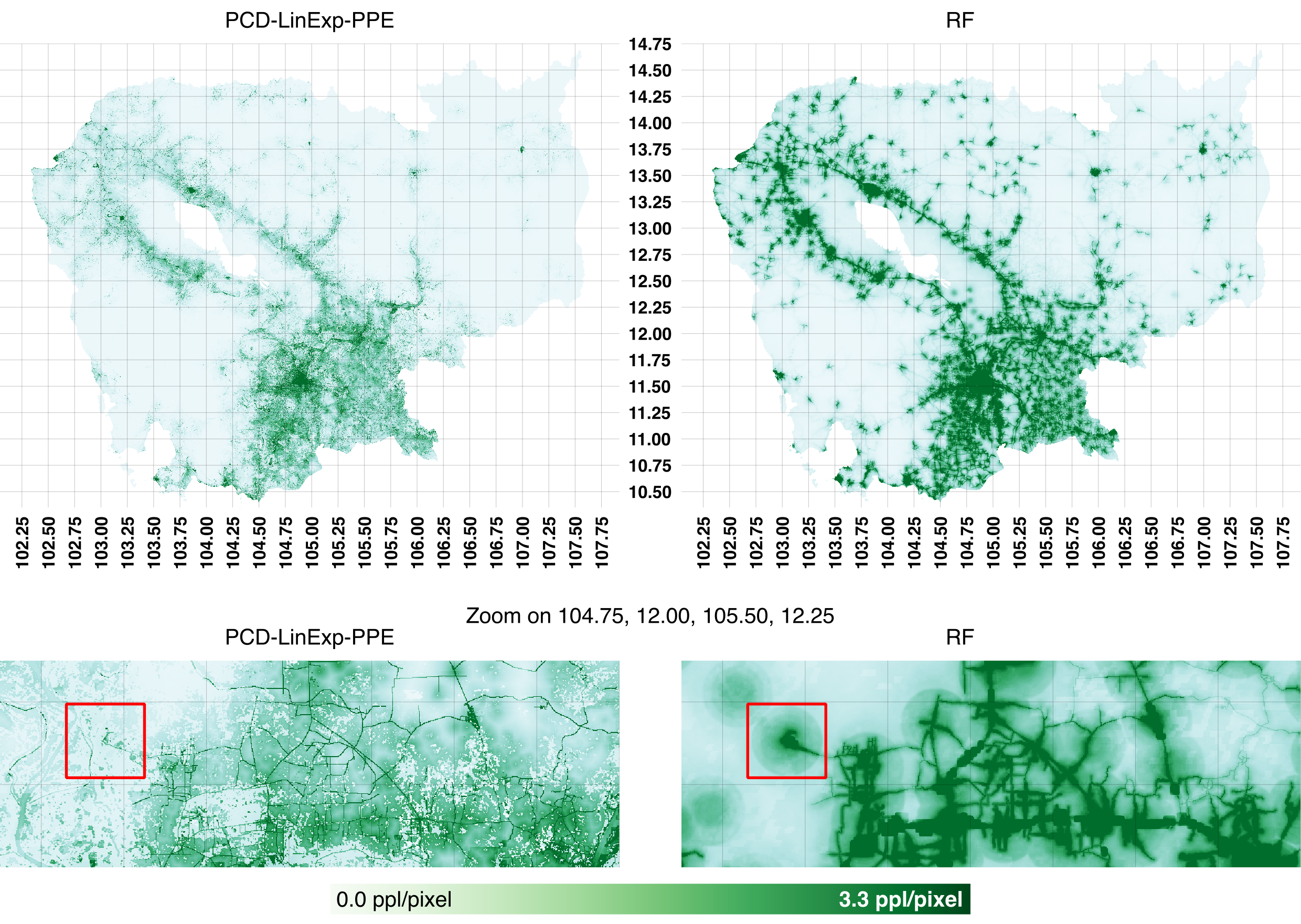}
	\caption{Cambodian population maps obtained by PCD-LinExp-PPE (a specialization of PCD using the PCD-LinExp model type, introduced in a later section) and RF. The bottom maps are a zoom of a specific, moderately populated region of Cambodia. The red box highlights a region where RF produces seemingly artificial results: it creates a circle around a (non-displayed) hospital, and saturates near the road network.}
	\label{map1}
\end{figure}

These models show various degrees of usability and accuracy. Among the methods cited above, RF \cite{rf} (for Random Forest, the family of algorithms on which the method is based) gives the best results using remotely sensed and ancillary data. However, the results of RF are seemingly artificial. The right panel of Fig~\ref*{map1} shows a fine-grained population mapping of Cambodia generated with the RF model from \cite{rf}. Geometrical shapes clearly emerge from the population disaggregation. On the contrary, as depicted on the left panel of Fig~\ref*{map1}, our PCD (for Pure Census Disagregation) method generates maps with a natural distribution aspect.

All methods in the literature are converted into standard supervised regression learning tasks. Supervised regression learning aims to predict an output value associated with a particular input vector. In its standard form, the training set contains an individual output value for each input vector.
Unfortunately, the disaggregation problem does not directly fit into this supervised regression learning framework 
since the prediction function is not directly available for input vectors. In a disaggregation problem, the input consists of a partition of the training set (the pixels of each unit) and for each partition, the sum of the output values is constrained by the census count. This framework is exactly the one introduced as the aggregated output learning problem by \cite{musicant2007supervised}. Our PCD method conceives the formulation of the disaggregation problem as an aggregated output learning problem. PCD is able to train the model based on a much larger training set composed of pixels. This approach is parameterized by the error function that the user seeks to minimize.

PCD can be used with a large variety of model types. In this paper, we consider a specific model, arguably simplistic, that is called PCD-LinExp. The PCD-LinExp model can be written as:
$$f_{\bm{\theta}=(\bm{a},b,c)}(\bm{D}_{p_x, p_y}) = \max(0,\exp(\bm{a}^T\bm{D}_{p_x,p_y}+b)+c),$$
where $f$ is the predicted population count, the vector $\bm{a}$ and scalars $b$ and $c$ are parameters to estimate, and $\bm{D}_{p_x,p_y}$ is a set of sensed/ancillary covariates of a given pixel located at $p_x, p_y$.

PCD-LinExp is arguably simplistic. This is by design, for two reasons. First, it is interpretable: each available remotely sensed covariate is associated to a single parameter in the vector $\bm{a}$. Second, the aim of this paper is to show that the aggregate learning approach is particularly suitable and gives good results on this particular problem. Future research will focus on more complex models. Despite its simplicity, we actually show in the result section that, due to its comparatively small unadjusted error compared to other methods, PCD-LinExp can be used to predict population counts when census data are not available, something that is not possible with existing methods.

As a case study, we experiment it on Cambodia using various sets of remotely sensed/ancillary data.  While this is not the main focus of this paper, we shortly interpret the result of our method on Cambodia. Our main result, the disaggregated map of Cambodia, is depicted on the left panel of Fig~\ref*{map1}. This paper also discusses methodological issues raised by existing approaches. In particular, we demonstrate that the previously used error metrics are biased when available census data involves administrative units with highly heterogeneous surfaces and population densities. We propose alternative metrics that better reflect the accuracy properties that should be fulfilled by a sound disaggregation approach. We then present the results for Cambodia and compare methods using various error metrics, providing statistical evidence that PCD-LinExp generates the most accurate results.

\subsection{Notations and definitions}



A country is divided into administrative units (abbreviated as AUs), that are part of various administrative levels and form a hierarchy. For example, in Cambodia, the AUs at the finest level of the AU hierarchy are called \textit{communes}, which are grouped into \textit{districts}. In most countries, the census is conducted at the finest possible level of the AU hierarchy. For simplicity, throughout this paper, we refer to an AU at the finest administrative level as \textit{units}, and use the term \textit{superunit} for an AU at the coarser level.

\begin{figure}[ht]
	\centering
	\includegraphics[width=0.7\textwidth]{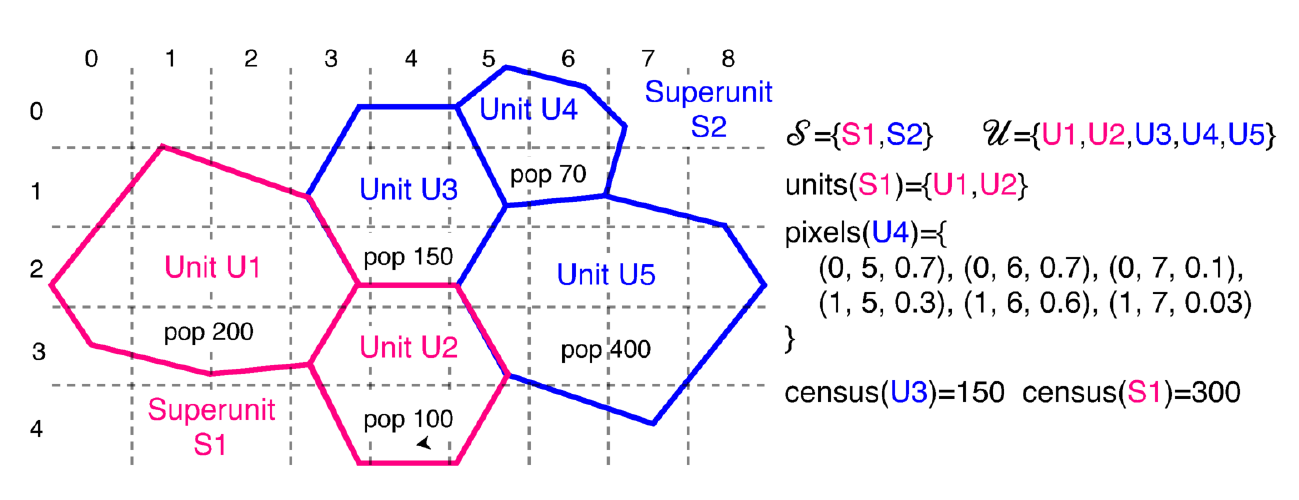}
	\caption{Notation example.}
	\label{fig-notations}
\end{figure}

\begin{definition}[Units, Superunits]
	Let $\mathcal{U}$ be the set of units and $\mathcal{S}$ the set of superunits. Let us denote by $\units(s)=\{u \mid u \in s \}$ the set of units in a superunit $s$. The set of pixels in a unit is $\pixels(u) = \{p \mid p \in u \}$ and by extension the set of pixels in a superunit is $\pixels(s)= \bigcup_{u \in \units(s)} \pixels(u)$. A pixel $p$ is a triple $(p_x,p_y,p_w)$, where $(p_x,p_y)$ is the position of the pixel and $p_w$ the weight of the pixel in the unit, which is the area of the pixel covered by the unit; $\surface(u)=\sum_{p \in \pixels(u)} p_w$ is the total surface of the unit u; $\surface(s)$ for superunits is defined similarly. We further define $\census(u)$ as the population count of unit $u$ and $\census(s)=\sum_{u \in \units(s)} \census(u)$ as the population count of superunit $s$. 
\end{definition}
An example illustrating the definitions is provided in Fig~\ref*{fig-notations}.

\subsection{Existing methods for disaggregation}
\label{other_work}

None of the existing approaches use the aggregate learning framework \cite{musicant2007supervised}. They use standard supervised learning approaches where the training set contains an individual output value for each input vector. The input data available for the disaggregation problem does not directly fit into this framework since the prediction value for each pixel is not available; an aggregate form is available, however (the census count for a unit). Existing disaggregation methods first transform the data to make them fit into the standard regression problem. Existing approaches are differentiated by their transformation method, by the error metric used to evaluate the quality of the learned function, and by the family of machine learning algorithm used to train the model.
\paragraph{Areal weighting} \cite{flowerdew1993developments,goodchild1993framework} is a straightforward method to redistribute the census data to a grid of pixels. The population predicted for each pixel is the sum of the contribution of each unit that contains the pixel. The contribution of a unit to a pixel $p$ is defined as $\prediction_u(p) := \frac{p_w}{\sum_{p' \in \pixels(u)} p'_w}\cdot \census(u).$ The Gridded Population of the World (GPW) dataset \cite{GPW} provides a raster (grid) of the population density/count at a fine resolution for the whole world. After census data is gathered, it is converted into a grid using an additional dataset on water bodies only. It is then a direct application of areal weighting, where each pixel weight $p_w$ is adjusted for the ratio of water bodies inside the pixel.

\paragraph{Dasymetric modeling} is an extension of areal weighting using ancillary data \cite{eicher2001dasymetric,wright1936method,mennis2003generating}. A weight is inferred from this ancillary data, which redistribute the population according to these weights. Land use/cover are often used as ancillary data \cite{DMOWSKA2014417,monmonier1984land,tian2005modeling}. Methods used to select the weights vary from purely heuristic choice of weights \cite{DMOWSKA2014417} to use of Expectation-Maximization or Logit regression \cite{Gallego2010}, and other  linear regression models \cite{BRIGGS2007451}. The methods below can all be seen as members of the Dasymetric modeling family, with the specificity that they all use more complex machine learning algorithm to predicts the weights accorded to each ancillary data type. They are also (mostly) agnostic on the type of the data and can handle multiple types at the same time, in a generic way, which is not the case for the methods presented above.

\paragraph{Random Forest (RF) method: pixel aggregation}
The methodology followed by \cite{rf} solves a 
regression problem in order to predict, for any pixel on a grid, the logarithm of the population count in this pixel. It is based on nonparametric machine learning techniques (RF) and uses observable covariates such as presence of hospitals, road networks, nightlight intensities, etc. 
Each covariate is mapped/projected on a grid of pixels (also denoted raster) spreading over the selected country. 
The resolution of this grid is decided by the user. In \cite{rf}, it is set at $8.33e^{-4}$ degrees, which 
approximately corresponds to ~100 m sided square pixels at the equator. A key contribution of \cite{rf} is the methodology to extract a training set used by the RF \cite{Breiman2001} machine learning algorithm. 

One data point is produced for each unit in the original training set. Each data point is a vector, containing a value (binary or floating point) for each covariate. These values are resulting from the pixel aggregation step, which varies with the type of covariate: for continuous covariates, the result of the aggregation of pixels is the mean of the covariate values of these pixels. For binary covariates, the mode (class majority) of the pixel's covariate values is taken as a summary for the unit. We thus obtain $|U|$ data points (remember that $|U| << $ the number of pixels), having each a fixed size, suitable for standard supervised machine learning techniques, such as random forests, neural networks, or SVMs. The training set is then used to learn a RF regression model.

Let us denote the training data as $\bm{D}$, a tensor in three dimensions, such that $\bm{D}_{x,y,f}$ gives the value of the covariate $f$ at position $x,y$. $\bm{D}_{x,y}$ gives the vector of all the covariate values at position $x,y$. The regression $RF(\bm{D}_{x,y})$ is thus able to predict a population density for a pixel with covariates values $\bm{D}_{x,y}$.

\paragraph{Corrected output of linear regression model (dasymetric redistribution)}
\label{section_dasymetric_redistribution}
The method used by \cite{rf} does not directly consider the output of $RF(\bm{D}_{x,y})$ as the predictor for the population density. Instead, it considers the output of the $RF$ for each pixel as a weighting layer used for subsequent dasymetric redistribution (also called pycnophylactic redistribution  \cite{tobler1979smooth}).

For each unit $u$, a correcting factor $w(u)$ is computed as 
$$w(u)=\frac{ \census(u)}{\sum_{p \in \pixels(u)} p_w\cdot RF(\bm{D}_{p_x,p_y})}.$$
The predicted density estimator for the pixel $p$ lying in unit
$u$ is then $$\prediction(p) = RF(\bm{D}_{p_x,p_y})\cdot w(u).$$

During the training of the model, this would, of course, lead to an error of zero. Therefore, the dasymetric redistribution is only applied on superunits. This part of the method obviously limits the usage of the method to situations where the census data is available, at least at a coarse resolution. We call the error \emph{adjusted} if it is dasymetrically redistributed, and \emph{unadjusted} otherwise.

\paragraph{Learning from heuristic, using satellite images and  deep neural networks}
\cite{robinson2017deep} relax the problem in a different way; they use an external census grid, produced in a heuristic way \cite{USGrid}, as ground truth. More precisely, they train their model based on a grid made from the 2000 US census, and validate it by measuring the prediction accuracy on the 2010 US census. This ``relaxation'' of the initial problem allows using standard supervised machine learning techniques. As these census grids are extrapolated from the official census, they suffer from higher errors, making the dataset noisier. However, compared to the Pixel Aggregation relaxation, it creates a much larger dataset, with one data point per pixel rather than one point per unit.
In their paper, Robinson et al. do not use ancillary data. They use satellite images in the visible and invisible spectrum. They use a Convolutionnal Neural Network model (abbreviated CNN, see \cite{lecun1998gradient}), which is a modified VGG-A neural net \cite{simonyan2014very}. The neural net is given a window of pixels of size $74\times 74$ (each pixel being approximately $15m\times 15m$), and predicts a (single) class for this set of pixels. It is a classification task, each class representing a given population range, obtained after binning the US censuses. The method uses a different type of data, among which some are not available in most countries (census grids). Hence, the interest in comparing this method with others is limited, and is thus not included here. Another method, LL \cite{ll} use the same preprocessing method (using a nearest neighbor method to produce the heuristic grid). LL is a prediction method that converts all covariates into a grid with pixels of size 250m$\times$250m (approx. 0.0625 km$^2$), and then assigns a ground truth for the population count of each pixel. The latter is proxied with the mean population density of the \textit{unit} they belong to. As for the previous method, we do not compare our method to LL. The data types are different (they only rely on satellite images) and the results are too coarse (both RF and our method uses pixels that are 6.25 times smaller in surface). In the original paper \cite{ll}, the LL method produces good results but only after applying the dasymetric redistribution (as with RF).

\section{Data preparation}
In order to compare our PCD-LinExp method with existing ones, various data sets have been collected for Cambodia. These include both data sensed by satellites
, and data provided by the Cambodian administration and by NGOs
. The description of the dataset, along with the preprocessing done, is available in the additional resources of this paper. Both the data and the preprocessing are similar to the ones in the article describing the RF method \cite{rf}.

In our experiments, we select a resolution of $8.33e^{-4}$ degrees, which approximately corresponds to squares of 100m sides at the equator. We then obtain a stack of rasters, which can be represented as a tensor in three dimensions. We call this tensor $\bm{D}$, with $\bm{D}_{x,y,f}$ being the pixel of the feature $f$ at position $x,y$. In the case of Cambodia, the dimension of this tensor is $6617 \times 5468 \times 42$. We use $\bm{D}_{x,y}$ as notation for the vector of all the covariates at position $x,y$. One difference between our method to prepare data and the one in \cite{rf} is that our data are standardized. In order to reduce numerical errors and make covariates equally important at the beginning of the computation, each feature is centered (resulting covariates have their mean equal to zero) and transformed such that they have a unit variance.


\section{Evaluation}
\subsection{Error metrics}
\label{error_metrics}
Assuming the finest ground truth we have is the administrative unit of the census, all errors metrics are in the form $e(\census, \prediction, \surface, \mathcal{U})$, where $\census$, $\prediction$ and $\surface$ are functions $\mathcal{U} \rightarrow \mathbb{R}^+$ that give the effective and predicted population of a unit and its surface, respectively.

Existing methods\cite{rf,ll,robinson2017deep} mostly use the RMSE criterion and its variants to estimate the prediction error. In our context, the RMSE is defined as:

$$e_{RMSE}=\sqrt{\frac{\sum_{u \in \mathcal{U}} \left ( \prediction(u) - \census(u) \right ) ^2}{|\mathcal{U}|}}.$$

RMSE is biased towards highly populated, low-surface administrative units. An analysis of the data on Cambodia shows that there are smaller-surface units than high-surface ones, and that the small ones tend to be more populated. In most countries, the division of the country in administrative units is such that the number of inhabitants in each unit is not too large.


Since there are more small regions than large ones, methods that better predict these regions have better scores. Said differently, $e_{RMSE}$ overemphasizes errors on small, highly populated regions. We are aware that this is not always a desirable property. Users might be more interested in accurately disaggregating high-surface administrative units than smaller units with already more precise population count.

\begin{figure}[ht!]
	\centering
	\includegraphics[width=0.6\textwidth]{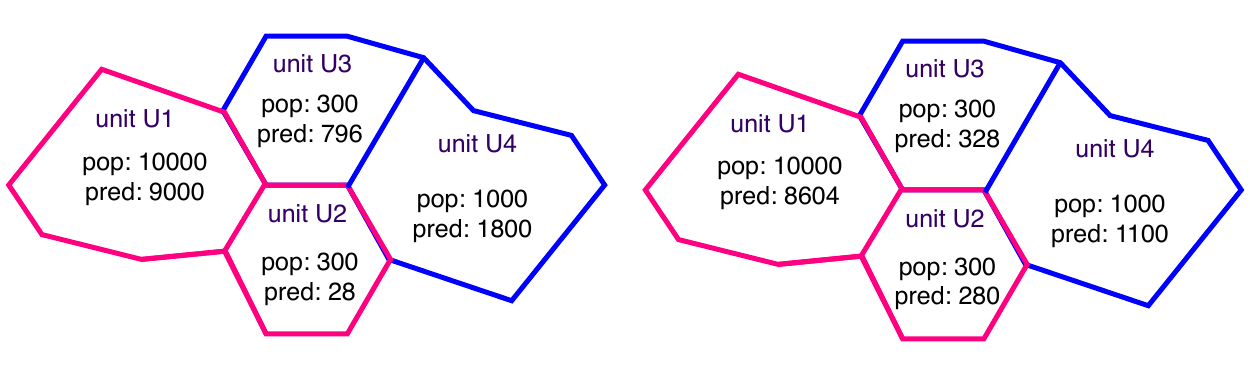}
	\caption{Example of administrative units with equal RMSE. In this diagram, for simplicity, \textit{pop} and \textit{pred} are respectively abbreviations for $\census$ and $\prediction$.}
	\label{error_demo_surface}
\end{figure}

Fig~\ref*{error_demo_surface} provides an example of two sets of units, with the same population, but with different predictions. Both of them have $e_{RMSE}=700$.  As can be seen, the example on the left panel makes very important errors on units U3 (error = 496) and U2 (272), where actual population counts are relatively small. However, these errors do not contribute a lot to the RMSE and are instead absorbed by the error of unit U1 (1000), which is relatively small compared to its population, but large in absolute value. On the contrary, the example on the right panel of Fig~\ref*{error_demo_surface} gives a more balanced error distribution. In absolute value the error is greater in U1. However, compared to the actual population counts, they are smaller in each administrative unit.

We postulate that a sound error metric for the disaggregation problem should be expressed in per-pixel units, therefore being independent of the surface of each unit and be based on relative population deviations rather than on absolute population counts.

We introduce the Per-Pixel-Error (PPE):
$$e_{PPE} = \frac{1}{\sum_{u\in \mathcal{U}} \surface(u)}\sum_{u\in \mathcal{U}} \surface(u) \cdot \frac{|\prediction(u)-\census(u)|}{\census(u)},$$
which is a surface-weighted average of relative deviations from the effective population (i.e. a Weighted Mean Absolute Percentage error, with surfaces as weights). PPE is thus independent from the size of the unit. Moreover, the use of relative deviations allows to effectively compare less-populated units to more-populated ones. In the examples from Fig~\ref*{error_demo_surface}, considering that all units have the same surface, the left example has a PPE of $86.5\%$ while the right one has a PPE of $10\%$. Remember that they have exactly the same RMSE (700).

A squared variant of PPE, the Per-Pixel Squared Error (PPSE), is defined as:
$$e_{PPSE} = \frac{1}{\sum_{u\in \mathcal{U}} \surface(u)}\sum_{u\in \mathcal{U}} \surface(u) \cdot \frac{(\prediction(u)-\census(u))^2}{\census(u)},$$
(which is a Weighted Root Mean Squared Error metric weighed by surface) which overweights greater error ratios. 

In the quantitative analysis below, we conduct our experiments using PPE and PPSE as error metrics. We also use RMSE so as to compare our results with previous works. MAE was used in previous works, but it suffers from the same shortcomings as RMSE and does not provide additional relevant information, and as such is omitted in our experiments.

\subsection{Comment on dasymetric redistribution}
\label{comment_dasy}
As discussed previously, dasymetric redistribution is a commonly used processing method to rescale (ex-post) the population in a given surface using a weight map. This method is used by RF \cite{rf}. The main argument justifying the use of dasymetric redistribution is the full exploitation of the available information included in the population census. If a disaggregation method predicts that there are 3,500 people in a given AU, and that the census reports an actual population of 4,500, dasymetric redistribution consists of multiplying all pixels within the AU by $\frac{4500}{3500}$, thus reducing the visible error.

While the method seems intuitive and duly justified, it actually increases the prediction error in some cases. This is not only due to the fact that the population census is, like all types of databases, subject to errors and various noises. It is also due to the fact that large local errors in a given region are \textit{transferred} to the whole region.

As an example, let us consider the case where we have two units composed each of one pixel. These units are grouped into a single superunit. Table~\ref*{example_dasy_distrib} shows (on the left) the actual and predicted population counts of these units, along with the adjusted prediction after dasymetric redistribution on the superunit. The total population of the superunit is $200$, but the predicted population is $160$. The correcting factor to be applied is $\frac{200}{160} = 1.25$. The right part of Table~\ref*{example_dasy_distrib} shows the error metrics applied before and after the dasymetric redistribution. In our example, both metrics increase after the redistribution. The reason is the error made originally on unit B. The low prediction forces the redistribution to increase the global count, dramatically increasing the error on unit A.

\begin{table}[h!]
	\caption{Example for dasymetric redistribution}
	\centering
	\begin{tabular}{|l|c|c|c|}
		\hline
		Unit & Real pop.& Pred. pop. & Adj. pop.\\
		\hline
		A & 100 & 120 & 150\\
		B & 100 & 40 & 50\\
		\hline
	\end{tabular}\hspace{1cm}\begin{tabular}{|l|c|c|}
	\hline
	Metric & Unadjusted & Adjusted \\
	\hline
	RMSE & $\sqrt{2000}$ & $\sqrt{2500}$ \\
	PPE & $40\%$ & $50\%$ \\
	\hline
	\end{tabular}
	\label{example_dasy_distrib}
\end{table}

Another drawback of dasymetric redistribution is that it breaks the dichotomy between inputs and outputs in the validation stage. Indeed, the dasymetric redistribution is applied to superunits (it would lead to an error of zero if applied to units). This implies that part of the prediction truth values are used to correct the output, before prediction errors are computed. We leave as an open question the relevance of using dasymetric redistribution. In practice, we notice that it sometimes increases the error in our experiments. Dasymetric redistribution also tends to make comparisons between methods difficult, as it \textit{squeezes} the original errors. To allow comparison with previous works, we produce both the unadjusted and adjusted (by dasymetric redistribution) results in our experiments.

\section{Pure Census Disaggregation (PCD)}

Existing methods require preprocessing the data before estimating the model. We propose to formalize the disaggregation problem in the context of aggregated output learning \cite{musicant2007supervised}. We call this formulation PCD. Aggregated output learning is a family of methods where the response information (the value to be predicted) about each data point is unknown. However, for specific subsets of data points, the sum of their responses is known. The disaggregation problem can be converted into an aggregated output learning problem: each pixel is a data point, and the subsets of data points for which we have the sum of the responses are AUs' pixels (the aggregated responses are the census counts).

In our context, solving the disaggregation problem consists of learning a function $f$ that, when applied and aggregated over all pixels of an AU, predicts the census count of this AU with accuracy. More precisely, we attempt to find the parameters $\bm{\theta}$ of a model $f_{\bm{\theta}}$ such that a given error metric ($e_{PPE}$, $e_{PPSE}$, $e_{RMSE}$, etc.) is minimized:

$$\argmin_{\bm{\theta}} error(\census, \prediction_{\bm{\theta},\bm{D}}, \surface, \mathcal{U}),$$
where $\prediction_{\bm{\theta},\bm{D}}$ is a function $\mathcal{U} \rightarrow \Re^+$ in the following form:

$$\prediction_{\bm{\theta},\bm{D}}(u) = \sum_{p\in \pixels(u)} p_w \cdot f_{\bm{\theta}}(\bm{D}_{p_x, p_y}).$$

Unlike the alternative techniques discussed previously, the error function is plugged into the PCD problem being globally minimized by gradient descent, in order to find $\bm{\theta}$. Other techniques are applicable to solve aggregated output learning problems \cite{musicant2007supervised} and can be applied to PCD. The prediction function for a unit is not based on a summary; it is composed of individual and unsupervised predictions for all pixels. 


The method is semi-supervised: while information is given to the method about administrative units of the census, no indication about the population is available at the individual pixel level. We use the name PCD-LinExp to indicate the usage of LinExp in the PCD framework. 

\subsubsection{An interpretable model}
We decide to focus on a regression model that we call LinExp. By design, it is easily interpretable. Remember the model can be written as: 

$$f_{(\bm{a},b,c)}(\bm{D}_{p_x, p_y}) = \max(0,\exp(\bm{a}^T\bm{D}_{p_x,p_y}+b)+c),$$
where $\bm{a}$ is a vector containing one parameter for each covariate, whose scalar product with the vector of covariate at position ($p_x,p_y$) is given to the $\exp$ function; $b$ and $c$ are two scalars allowing correcting the output. The output is then passed into a function $\max(0,\bullet)$ (a ReLU \cite{hahnloser2000digital}, here used outside its traditional scope) to ensure the output is positive. The presence of the exp function is motivated by the distribution of the census.

As the covariates in the tensor $\bm{D}$ have been standardized (they are centered and have unit variance), the vector $\bm{a}$ is directly interpretable as the importance of each covariate: 
\begin{itemize}
	\item a high absolute value indicates that the covariate contributes highly to the prediction, while a low absolute value indicates its contribution is small;
	\item positive values indicate that each increase of the covariate increases the predicted population, while negative values do the opposite.
\end{itemize}

The model is convex; however, when plugged into an error function such as $e_{RMSE}$ or $e_{PPE}$, it becomes non-convex, mainly due to the sum of absolute (or squared) values of exponentials. Using a (sub-)gradient descent algorithm minimizing the error function on the training set is then a simple yet effective choice to find good values for parameters $\bm{a},b,c$.

The function to be optimized for PCD-LinExp, given that $e_{PPE}$ is used, is
\begin{multline*}
	\argmin_{\bm{a},b,c} \frac{1}{\sum_{u\in \mathcal{U}} \surface(u)} \sum_{u\in \mathcal{U}} \\\left( \surface(u) \cdot \frac{|(\sum_{p\in \pixels(u)} p_w \cdot \max(0,\exp(\bm{a}^T\bm{D}_{p_x,p_y}+b)+c))-\census(u)|}{\census(u)}\right).
\end{multline*}

\subsubsection{Optimization} Due to the presence of the $\exp$ function in the model, it is quite sensitive to sudden erroneous increases in the vector $\bm{a}$, which can make the output explode and even overflow. Hence, relatively small learning rates are used, and we choose to avoid the use of stochastic gradient descent (SGD): the whole dataset is computed at each iteration. This is allowed by the relatively small size of our dataset; bigger datasets (not fitting entirely in memory) would require using SGD.


Optimization is done using the Adam algorithm \cite{DBLP:journals/corr/KingmaB14}. Adam, for \textit{Ada}ptive \textit{m}oment estimation, is a variation of the standard Stochastic Gradient Descent algorithm; it uses on first-order gradients and estimations of the first and second moments of these gradients to regularize learning steps.

We use learning rates of $0.01$ for vectors $a$ and $b$, and of $0.001$ for vector $c$. These values result from manual tuning. We run $1000$ iterations of Adam before finishing and returning the results. Experiments show that convergence is most of the time reached before $\sim500$ iterations.

\subsubsection{Initialization} Concerning initialization, we attempt to have at the first iterations a small value for $\bm{a}^T\bm{x}+b$, as a large value may produce very important gradients, which is not desired. Experiments have shown that having $\bm{a}=\bm{0}$ and a negative value for $b$ (we chose $-4$) works well in practice. Indeed, it greatly underestimates the population during the first few iterations of the gradient descent.

\subsubsection{Experimental Setup}
\label{experimental_setup}
In order to evaluate correctly the different methods (PCD-LinExp, RF) presented earlier, we implemented a common test framework in Python. The PyTorch library \cite{paszke2017automatic} was chosen to create the PCD-LinExp models. The RF implementation relies on the library provided in the first publication of the method (i.e. the \texttt{randomforests} library of the R language). The new RF implementation has been tested in similar conditions than in the original paper and gives similar results (in fact, we obtain better results, which is probably due to small differences in the data).

We used standard 10-fold cross-validation to test methods. Cambodia is divided into 193 districts, which are themselves divided into 1621 communes. As we have to test the dasymetric redistribution, we need to keep communes inside the same district into the same fold. We thus obtain folds with 19 districts (three folds have 20 districts). As far as the cross-validation works is concerned, a fold is used as test set, and the other nine folds are used as training set. The fold used to test the models is then changed. We then obtain 10 measures of the generalization of the methods, from which we take the mean. The measures chosen are the Per-Pixel-Error (PPE), the Per-Pixel-Squared-Error (PPSE), and the Root Mean Square Error (RMSE). 

\subsubsection{RF-adj}

In order to produce more accurate comparisons between the different methods, we introduce a small variation of the RF method, that we call RF-adjusted (abbreviated as RF-adj). The motivation behind RF-Adj is that RF tends to over-predict population by a fair margin, as it is not constructed to run on unadjusted errors (it was originally described with dasymetric redistribution included in the method \cite{rf}).

RF-Adj is a simple adjustment that minimizes the error on the training set. Given a trained $RF$ function, the $RF_{ajd}$ function is defined as $RF_{adj}(\bm{D}_{p_x,p_y}) = c \cdot RF(\bm{D}_{p_x,p_y}),$ where $c$ is the mean factor of error for each unit in the training set:
$$c = \frac{\sum_{u\in \mathcal{U}_{\text{training}}} \frac{\prediction_{\text{RF}}(u)}{\census(u)}}{|\mathcal{U}_{\text{training}}|}.$$

Experiments show that RF-Adj generates smaller \textit{unadjusted} errors than RF.

\section{Results and discussion} 
Fig~\ref*{map1} (at the beginning of the paper) compares the population maps generated by the PCD-LinExp-PPE and RF methods, before dasymetric redistribution. Results obtained by PCD-LinExp-RMSE and RF-Adj are not included as there are visually similar to PCD-LinExp-PPE and RF, respectively. The two methods identify similar regions of high population density, notably near the capital Phnom Penh (at the south-east of the country) and around the lake Tonlé Sap, in the center of the country. Visually, PCD-LinExp-PPE produces a seemingly more detailed map than RF, with less ``saturation''. On the one hand, PCD-LinExp-PPE highlights the role of the road infrastructure. On the other hand, RF produces circles of high population density around hospitals and roads.

\begin{figure}[ht!]
	\centering
	\includegraphics[width=0.7\textwidth]{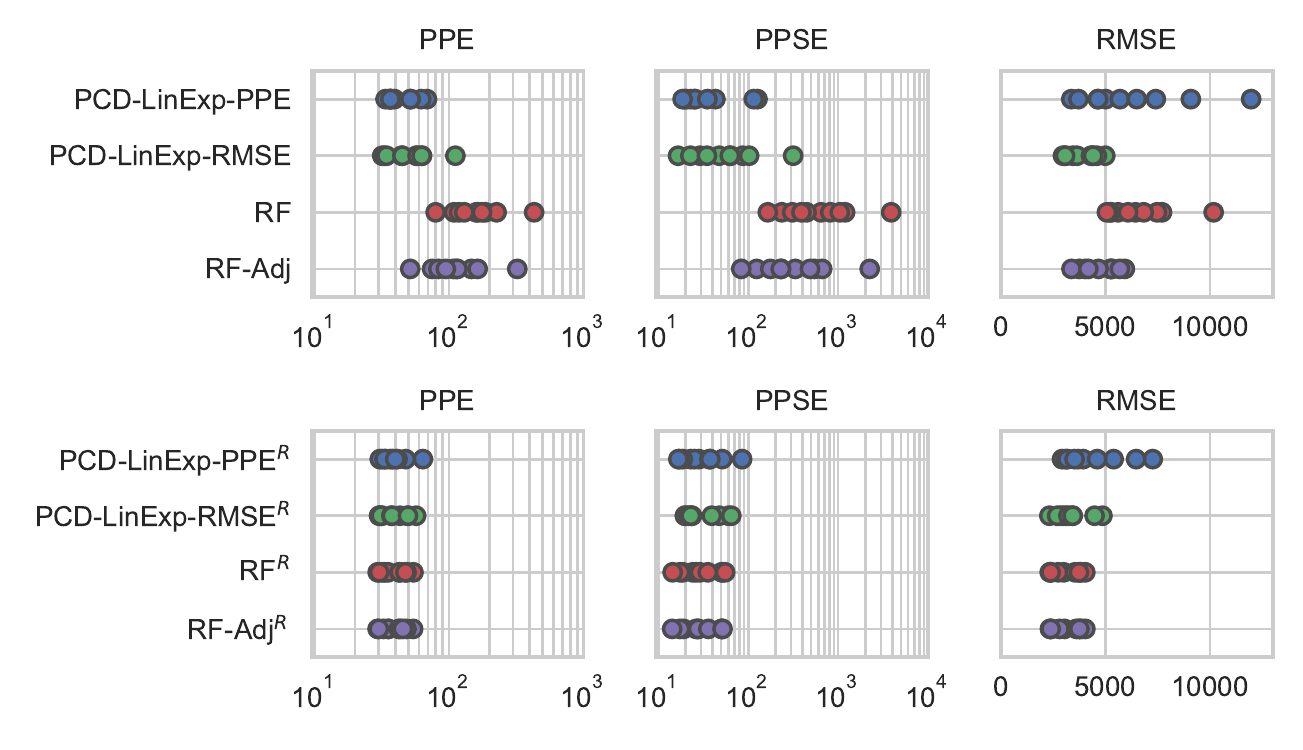}
	\caption{Scatter plot of the error metrics for each model and for each fold. While it is difficult to compare results for the redistributed errors, PCD-LinExp-PPE and PCD-LinExp-RMSE give the best result on unadjusted errors. The methods PCD-LinExp-PPE$^R$, PCD-LinExp-RMSE$^R$, RF$^R$ and RF-Adj$^R$ are redistributed counterparts of the original methods.}
	\label{error_boxplot}
\end{figure}

Table~\ref*{table_error_metric_full} shows the error metrics of the methods described in our experimental setup. Box plots containing the data of the individual folds are depicted in Fig~\ref*{error_boxplot}. Running a paired t-test (null hypothesis mean(A-B) = 0, and a p-value limit of $0.05$) on the fold errors reveals that LinExp-PPE and LinExp-RMSE significantly less PPE/PPSE than RF and RF-Adj on unadjusted metrics. LinExp-RMSE is significantly better than all the other methods on unadjusted RMSE. On adjusted metrics, no method is significantly better than another, but LinExp-PPE that is dominated by all method on the adjusted RMSE metric.


PCD-LinExp-PPE and PCD-LinExp-RMSE better perform when using the unadjusted PPE/PPSE metric; PCD-LinExp-RMSE generates better results for the unadjusted RMSE metric. The PCD-LinExp unadjusted outcomes are comparable to the adjusted outcomes obtained after dasymetric redistribution, which strengthens our claim that PCD-LinExp can be used to predict the population where census data are not available.

\begin{table}[!ht]
	\caption{{\bf Error metrics}}
		\centering
		\begin{tabular}{|l+r|r|r+r|r|r|}
			\hline
			{\bf Method} & \multicolumn{3}{c+}{\bf Unadjusted} & \multicolumn{3}{c|}{\bf Adjusted}\\ 
			& PPE & PPSE & RMSE & PPE & PPSE & RMSE \\ \thickhline
			Areal weighting & N/A & N/A & N/A & $87.90\ \%$ & $215.46\ \%$ & $6777.71$\\ \hline
			PCD-LinExp-PPE & $45.43\ \%$ & $45.46\ \%$ & $6202.23$ & $39.75\ \%$ & $34.19\ \%$ & $4407.64$\\\hline
			PCD-LinExp-RMSE & $54.20\ \%$ & $78.35\ \%$ & $3889.35$ & $39.84\ \%$ & $34.45\ \%$ & $3335.68$\\\hline
			RF & $178.07\ \%$ & $914.97\ \%$ & $6584.09$ & $39.74\ \%$ & $32.44\ \%$ & $3175.68$\\\hline
			RF-Adj & $127.50\ \%$ & $512.72\ \%$ & $4742.39$ & $39.68\ \%$ & $32.24\ \%$ & $3199.69$\\\hline
	\end{tabular}
	\label{table_error_metric_full}
\end{table}

\begin{figure}[ht!]
	\centering
	\includegraphics[width=0.7\textwidth]{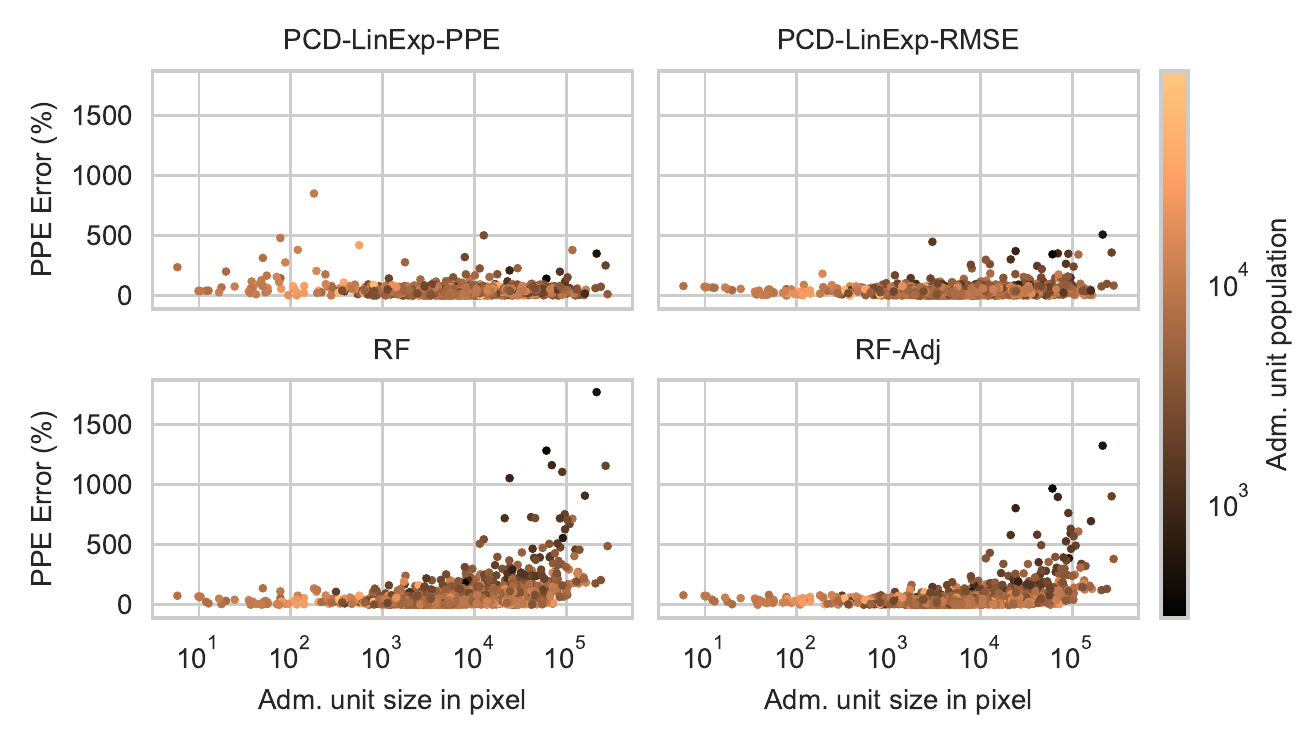}
	\caption{Error for each unit for all models.}
	\label{error_surface}
\end{figure}

Fig~\ref*{error_surface} shows the unadjusted PPE error obtained for each unit as a function of the surface of the unit. PCD-LinExp-PPE and PCD-LinExp-RMSE generate smaller errors. The errors obtained under PCD-LinExp-PPE are globally independent of the unit surface. This is because the PPE metric reduces the influence of the surface. As expected, other methods that do not account for the surface  generate greater errors for less-populated, high-surface units. Smaller errors are obtained for highly populated, small-surface units as this would greatly impact the RMSE.

%

Overall, all results presented in this section indicate that PCD-LinExp-PPE and PCD-LinExp-RMSE generate better results than RF and RF-Adj, at least for predicting the population (i.e. on unadjusted errors). The comparison for population disaggregation (using adjusted errors) is more difficult, as all experiments indicate ties. We explained previously that using dasymetric redistribution breaks the training set/test set dichotomy, and thus should be avoided. However, this demonstrates that using the aggregate learning framework with a very simple model in this problem outperforms/is on par (depending on the metric) with more complex models not using aggregate learning.

\subsubsection{Acknowledgments}
Computational resources have been provided by the supercomputing facilities of the Université catholique de Louvain (CISM/UCL) and the Consortium des Équipements de Calcul Intensif en Fédération Wallonie Bruxelles (CÉCI) funded by the Fond de la Recherche Scientifique de Belgique (F.R.S.-FNRS) under convention 2.5020.11. We would like to thank Pavel Demin and the CP3 group that shared with us part of their reserved resources. The second and third authors acknowledge financial support from the ARC convention on ``New approaches to understanding and modeling global migration trends'' (convention 18/23-091).

\bibliographystyle{splncs04}

\end{document}